# Learning Multi-Class Neural-Network Models from Electroencephalograms

Vitaly Schetinin[1], Joachim Schult[2], Burkhart Scheidt[2], and Valery Kuriakin[3]

[1]Department of Computer Science, Harrison Building, University of Exeter, EX4 4QF, UK
V.Schetinin@ex.ac.uk
[2]Friedrich-Schiller-University of Jena, Ernst-Abbe-Platz 4, 07740 Jena, Germany
Joachim_Schult@web.de
[3]Intel Russian Research Center, N. Novgorod, Russia

**Abstract.** We describe a new algorithm for learning multi-class neural-network models from large-scale clinical electroencephalograms (EEGs). This algorithm trains hidden neurons separately to classify all the pairs of classes. To find best pairwise classifiers, our algorithm searches for input variables which are relevant to the classification problem. Despite patient variability and heavily overlapping classes, a 16-class model learnt from EEGs of 65 sleeping newborns correctly classified 80.8% of the training and 80.1% of the testing examples. Additionally, the neural-network model provides a probabilistic interpretation of decisions.

## 1  Introduction

Learning classification models from electroencephalograms (EEGs) is still a complex problem [1] - [7] because of the following problems: first, the EEGs are strongly non-stationary signals which depend on an individual Background Brain Activity (BBA) of patients; second, the EEGs are corrupted by noise and muscular artifacts; third, a given set of EEG features may contain features which are irrelevant to the classification problem and may seriously hurt the classification results and fourth, the clinical EEGs are the large-scale data which are recorded during several hours and for this reason the learning time becomes to be crucial.

In general, multi-class problems can be solved by using one-against-all binary classification techniques [8]. However, a natural way to induce multi-class concepts from real data is to use Decision Tree (DT) techniques [9] - [12] which exploit a greedy heuristic or hill-climbing strategy to find out input variables which efficiently split the training data into classes.

To induce linear concepts, multivariate or oblique DTs have been suggested which exploit the threshold logical units or so-called perceptions [13] - [16]. Such multivariate DTs known also as Linear Machines (LM) are able to classify linearly separable examples. Using the algorithms [8], [13] - [15], the LMs can also learn to classify non-linearly separable examples. However, such DT methods applied for inducing multi-scale problems from real large-scale data become impractical due to large computations [15], [16].

Another approach to multiple classification is based on pairwise classification [17]. A basic ides behind this method is to transform a $q$-class problem into $q(q - 1)/2$ binary problems, one for each pair of classes. In this case the binary decision problems are presented by fewer training examples and the decision boundaries may be considerably simpler than in the case of one-against-all binary classification.

In this paper we describe a new algorithm for learning multi-class neural-network models from large-scale clinical EEGs. This algorithm trains hidden neurons separately to classify all the pairs of classes. To find best pairwise classifiers, our algorithm searches for input variables which are relevant to the classification problem. Additionally, the neural-network model provides a probabilistic interpretation of decisions.

In the next section we define the classification problem and describe the EEG data. In section 3 we describe the neural-network model and algorithm for learning pairwise classification. In section 4 we compare our technique with the standard data mining techniques on the clinical EEGs, and finally we discuss the results.

## 2   A Classification Problem

In order to recognize some brain pathologies of newborns whose prenatal age range between 35 and 51 weeks, clinicians analyze their EEGs recorded during sleeping and then evaluate a EEG-based index of brain maturation [4] - [7]. So, in the pathological cases, the EEG index does not match to the prenatal maturation.

Following [6], [7] we can use the EEGs recorded from the healthy newborns and define this problem as a multi-class one, i.e., a 16-class concept. Then all the EEGs of healthy newborns should be classified properly to their prenatal ages, but the pathological cases should not.

To build up such a multi-class concept, we used the 65 EEGs of healthy newborns recorded via the standard electrodes C3 and C4. Then, following [4], [5], [6], these records were segmented and represented by 72 spectral and statistical features calculated on a 10-sec segment into 6 frequency bands such as sub-delta (0-1.5 Hz), delta (1.5-3.5 Hz), theta (3.5-7.5 Hz), alpha (7.5-13.5 Hz), beta 1 (13.5-19.5 Hz), and beta 2 (19.5-25 Hz). Additional features were calculated as the spectral variances. EEG-viewer has manually deleted the artifacts from these EEGs and then assigned normal segments to the 16 classes correspondingly with age of newborns. Total sum of the labeled EEG segments was 59069.

Analyzing the EEGs, we can see how heavily these depend on the BBA of newborns. Formally, we can define the BBA as a sum of spectral powers over all the frequency bands. As an example, Fig. 1 depicts the BBA calculated for two newborns aged 49 weeks. We can see that the BBA depicted as a dark line chaotically varies during sleeping and causes the variations of the EEG which significantly alter the class boundaries.

Clearly, we can beforehand calculate and then subtract the BBA from all the EEG features. Using this pre-processing technique we can remove the chaotic oscillations from the EEGs and expect improving the classification accuracy.

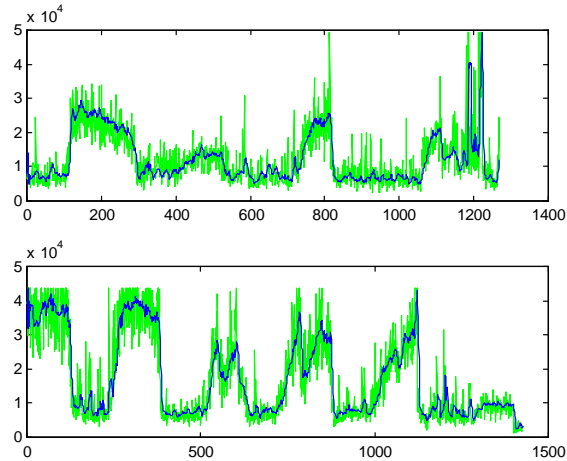

**Fig. 1.** EEG segments of two sleeping newborns aged 49 weeks

Below we describe a neural-network technique we developed to learn multi-class concept from the EEGs.

## 3 The Neural-Network Model and Learning Algorithm

The idea behind our method of multiple classification is to separately train the hidden neurons of neural network and combine them to approximate the dividing surfaces. These hidden neurons learn to divide the examples from each pair of classes. For $q$ classes, therefore, we need to learn $q(q-1)/2$ binary classifiers.

The hidden neurons that deal with one class are combined into one group, so that the number of the groups corresponds to the number of the classes. The hidden neurons combined into one group approximate the dividing surfaces for the corresponding classes.

Let $f_{i/j}$ be a threshold activation function of hidden neuron which learns to divide the examples $x$ of $i$th and $j$th classes $\Omega_i$ and $\Omega_j$ respectively. The output $y$ of the hidden neuron is

$$y = f_{i/j}(x) = 1, \forall\, x \in \Omega_i, \text{ and } y = f_{i/j}(x) = -1, \forall\, x \in \Omega_j. \tag{1}$$

Assume $q = 3$ classification problem with overlapping classes $\Omega_1$, $\Omega_2$ and $\Omega_3$ centered into $C_1$, $C_2$, and $C_3$, as Fig. 2(a) depicts. The number of hidden neurons for this example is equal to 3. In the Fig. 2(a) lines $f_{1/2}$, $f_{1/3}$ and $f_{2/3}$ depict the hyperplanes of the hidden neurons trained to divide the examples of three pair of the classes which are (1) $\Omega_1$ and $\Omega_2$, (2) $\Omega_1$ and $\Omega_3$, and (3) $\Omega_2$ and $\Omega_3$.

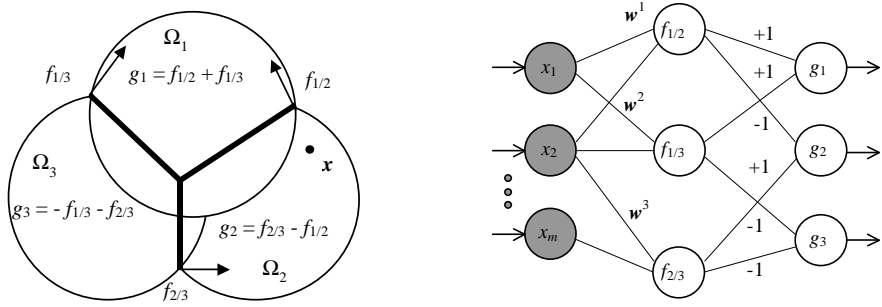

**Fig. 2.** The dividing surfaces (a) $g_1$, $g_2$, and $g_3$, and the neural network (b) for $q = 3$ classes

Combining these hidden neurons into $q = 3$ groups we built up the new hyperplanes $g_1$, $g_2$, and $g_3$. The first one, $g_1$, is a superposition of the hidden neurons $f_{1/2}$ and $f_{1/3}$, i.e., $g_1 = f_{1/2} + f_{1/3}$. The second and third hyperplanes are $g_2 = f_{2/3} - f_{1/2}$ and $g_3 = -f_{1/3} - f_{2/3}$ correspondently.

For hyperplane $g_1$, the outputs $f_{1/2}$ and $f_{1/3}$ are summed with the weights 1 because both give the positive outputs on the examples of the class $\Omega_1$. Correspondingly, for hyperplane $g_3$, the outputs $f_{1/3}$ and $f_{2/3}$ are summed with weights –1 because they give the negative outputs on the examples of the class $\Omega_3$.

Fig. 2(b) depicts for this example a neural network structure consisting of three hidden neurons $f_{1/2}$, $f_{1/3}$, and $f_{2/3}$ and three output neurons $g_1$, $g_2$, and $g_3$. The weight vectors of hidden neurons here are $\mathbf{w}^1$, $\mathbf{w}^2$, and $\mathbf{w}^3$. These hidden neurons are connected to the output neurons with weights equal to (+1, +1), (–1, +1) and (–1, –1), respectively. We can see that in general case for $q > 2$ classes, the neural network consists of $q$ output neurons $g_1, \ldots, g_q$ and $q(q-1)/2$ hidden neurons $f_{1/2}, \ldots, f_{i/j}, \ldots, f_{q-1/q}$, where $i < j = 2, \ldots, q$.

Each output neuron $g_i$ is connected to the $(q-1)$ hidden neurons which are partitioned into two groups. The first group consists of the hidden neurons $f_{i/k}$ for which $k > i$. The second group consists of the hidden neurons $f_{k/i}$ for which $k < i$. So, the weights of output neuron $g_i$ connected to the hidden neurons $f_{i/k}$ and $f_{k/i}$ are equal to + 1 and –1, respectively. As the EEG features may be irrelevant to the binary classification problems, for learning the hidden neurons we use a bottom up search strategy which selects features providing the best classification accuracy [11]. Below we discuss the experimental results of our neural-network technique applied to the real multi-class problem.

## 4 Experiments and Results

To learn the 16-class concept from the 65 EEG records, we used the neural network model described above. For training and testing we used 39399 and 19670 EEG segments, respectively. Correspondingly, for $q = 16$ class problem, the neural network includes $q(q-1)/2 = 120$ hidden neurons or binary classifiers with a threshold activation function (1).

The testing errors of binary classifiers varied from 0 to 15% as depicted in Fig. 3(a). The learnt classifiers exploit different sets of EEG features. The number of these features varies between 7 and 58 as depicted in Fig 3(b).

Our method has correctly classified 80.8% of the training and 80.1% of the testing examples taken from 65 EEG records. Summing all the segments belonging to one EEG record, we can improve the classification accuracy up to 89.2% and 87.7% of the 65 EEG records for training and testing, respectively.

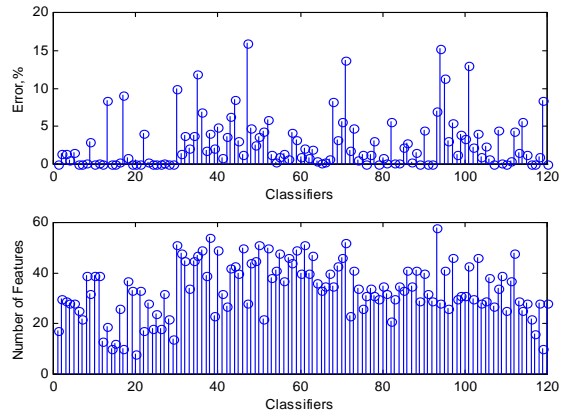

**Fig. 3.** The testing errors (a) and the number of features (b) for each of 120 binary classifiers

In Fig. 4 we depict the outputs of our model summed over all the testing EEG segments of two patients which belong to the second and third age groups, respectively. In both cases, most parts of the segments were correctly classified.

In addition, summing the outputs over all the testing EEG segments, we may provide a probabilistic interpretation of decisions. For example, we assign the patients to 2 and 3 classes with probabilities 0.92 and 0.58, respectively, calculated as parts of the correctly classified segments. These probabilities give us the additional information about the confidence of decisions.

We compared the performance of our neural network technique and the standard data mining techniques on the same EEG data. First, we tried to train a standard feedforward neural network consisted of $q = 16$ output neurons and a predefined number of hidden neurons and input nodes. The number of hidden neurons was defined between 5 and 20, and the number of the input nodes between 72 and 12 by using the standard principal component analysis.

Note that in our experiments we could not use more than 20 hidden neurons because even a fast Levenberg-Marquardt learning algorithm provided by MATLAB has required an enormous time-consuming computational effort. Because of a large number of the training examples and classes, we could not also use more than and 25% of the training data and, as a result, the trained neural network has correctly classified less than 60% of the testing examples.

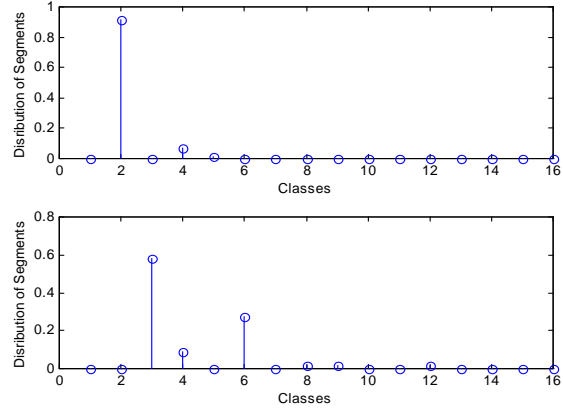

**Fig. 4.** Probabilistic interpretation of decisions for two patients

Second, we trained $q = 16$ binary classifiers to distinguish one class against others. We defined the same activation function for these classifiers and trained them on the whole data. However the classification accuracy was 72% of the testing examples.

Third, we induced a decision tree consisting of $(q - 1) = 15$ binary decision trees trained to split $(q - 1)$ subsets of the EEG data. That is, the first classifier learnt to divide the examples taken from classes $\Omega_1, \ldots, \Omega_8$ and $\Omega_9, \ldots, \Omega_{16}$. The second classifier learnt to divide the examples taken from classes $\Omega_1, \ldots, \Omega_4$ and $\Omega_5, \ldots, \Omega_8$, and so on. However the classification accuracy on the testing data was 65%.

## 5 Conclusion

For learning multi-class concepts from large-scale heavily overlapping EEG data, we developed a neural network technique and learning algorithm. Our neural network consists of hidden neurons which perform the binary classification for each pairs of classes. The hidden neurons are trained separately and then their outputs are combined in order to perform the multiple classification. This technique has been used to learn a 16-class concept from 65 EEG records represented by 72 features some of which were irrelevant.

Having compared our technique with the other classification methods on the same EEG data, we found out that it gives the better classification accuracy for an acceptable learning time. Thus, we conclude that the new technique we developed for learning multi-class neural network models performs on the clinical EEGs well. We believe that this technique may be also used to solve other large-scale multi-class problems presented many irrelevant features.

**Acknowledgments.** The research has been supported by the University of Jena (Germany). The authors are grateful to Frank Pasemann for enlightening discussions, Joachim Frenzel from the University of Jena for the clinical EEG records, and to Jonathan Fieldsend from the University of Exeter (UK) for useful comments.